\documentclass[10pt,twocolumn,letterpaper]{article}

\usepackage{cvpr}
\usepackage{times}
\usepackage{epsfig}
\usepackage{graphicx}
\usepackage{amsmath}
\usepackage{amssymb}
\usepackage{booktabs}
\usepackage{sidecap}
\usepackage{subcaption}
\usepackage{authblk}

\usepackage[pagebackref=true,breaklinks=true,letterpaper=true,colorlinks,bookmarks=false]{hyperref}

\cvprfinalcopy 

\def\cvprPaperID{4519} 

\begin{document}

\renewcommand*{\Authsep}{, \ }
\renewcommand\Authands{, \ }
\setlength{\affilsep}{0.5em}

\title{FPConv: Learning Local Flattening for Point Convolution}

\author[1,2]{Yiqun Lin}
\author[1,2]{Zizheng Yan}
\author[4]{Haibin Huang}
\author[2,3]{Dong Du}
\author[3]{Ligang Liu}
\author[1,2]{\\Shuguang Cui}
\author[1,2]{Xiaoguang Han\thanks{Corresponding Author: {\tt\small hanxiaoguang@cuhk.edu.cn}}}
\affil[1]{The Chinese University of Hong Kong, Shenzhen}
\affil[2]{Shenzhen Research Institute of Big Data}
\affil[3]{University of Science and Technology of China\ \ \ \ \ $^\text{4}$Kuaishou Technology}

\maketitle

\begin{abstract}

We introduce FPConv, a novel surface-style convolution operator designed for 3D point cloud analysis. Unlike previous methods, FPConv doesn't require transforming to intermediate representation like 3D grid or graph and directly works on surface geometry of point cloud.  To be more specific, for each point, FPConv performs a local flattening by automatically learning a weight map to softly project surrounding points onto a 2D grid. Regular 2D convolution can thus be applied for efficient feature learning. FPConv can be easily integrated into various network architectures for tasks like 3D object classification and 3D scene segmentation, and achieve comparable performance with existing volumetric-type convolutions. More importantly, our experiments also show that FPConv can be a complementary of volumetric convolutions and jointly training them can further boost overall performance into state-of-the-art results. Code is available at \href{https://github.com/lyqun/FPConv}{\tt\small https://github.com/lyqun/FPConv}

\end{abstract}

\section{Introduction}


With the rapid development of 3D scan devices, it is more and more easy to generate and access 3D data in the form of point clouds. This also brings the challenging of robust and efficient 3D point clouds analysis, which serves as an important component in many real world applications like robotics navigation, autonomous driving,  augmented reality applications and so on \cite{pomares2018ground, yue2018lidar, behl2017bounding, rambach2017poster}. 

Despite decades of development in 3D analysing technologies, it is still quite challenging to perform point cloud based semantic analysis, largely due to its sparse and unordered structure. Early methods \cite{charaniya2004supervised, chehata2009contribution, golovinskiy2009shape, martinovic20153d} utilized hand-crafted features with complex rules to tackle this problem. Such empirical human-designed features would suffer from limited performance in general scenes. Recently, with the explosive growth of machine learning and deep learning techniques, Deep Neural Network (CNN) based methods have been introduced into this task \cite{qi2017pointnet, qi2017pointnet++} and reveal promising improvements. However, both PointNet \cite{qi2017pointnet} and PointNet++ \cite{qi2017pointnet++} doesn't support convolution operation which is a key contributing factor in Convolutional Neural Network (CNN) for efficient local processing and handling large-scale data.

A straightforward extension of 2D CNN is treating 3D space as a volumetric grid and using 3D convolution for analysis \cite{wu20153d, riegler2017octnet}.  Although these approaches have achieved success in tasks like object classification and indoor semantic segmentation \cite{maturana2015voxnet, dai2017scannet}, they still have limitations like cubic growth rate of memory requirement and high computational cost, leading to insufficient analysis and low predication accuracy on large-scale scenes. Recently, \cite{wu2019pointconv, thomas2019kpconv} are proposed to approximate such volumetric convolutions with point-based convolution operations, which greatly improves the efficiency and preserves the output accuracy. However, these methods are still difficult to capture fine details on surface with relatively flat and thin structures. 



\begin{figure}[t]
\begin{center}
   \includegraphics[width=0.9\linewidth]{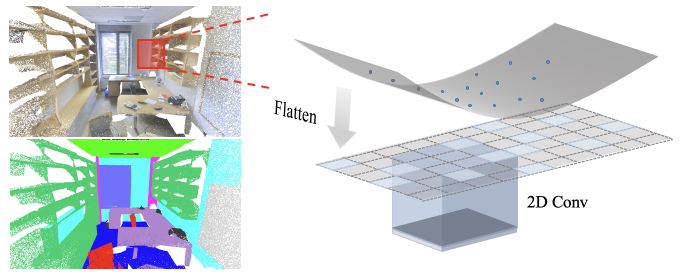}
\end{center}
\caption{Flattening Projection Convolution: flatten local patch onto a grid plane and then apply 2D convolutions.}
\label{fig:teaser}
\end{figure}

In reality, data captured by 3D sensors and LiDAR are usually sparse that points fall near scene surfaces and almost no points interior. Hence, surfaces are more natural and compact for 3D data representation. Towards this end,  works like \cite{defferrard2016convolutional, yi2017syncspeccnn} establish connections among points and apply graph convolutions in the corresponding spectral domain or focus on the surface represented by the graph \cite{simonovsky2017dynamic}, which are usually impractical to create and sensitive to local topological structures. 


More recently, \cite{tatarchenko2018tangent, pan2018convolutional, huang2019texturenet} are proposed to learn convolution on a specified 2D plane. Inspired by these pioneering works, we develop FPConv, a new convolution operation for point clouds. It works directly on local surface of geometry without any intermediate grid or graph representation. Similar to  \cite{tatarchenko2018tangent},  it works in projection-interpolation manner but more general and implicit. Our key observation is that projection and interpolation can be simplified into a single weight map learning process. Instead of explicitly projecting onto the tangent plane \cite{tatarchenko2018tangent} for convolution, FPConv learns how to diffuse convolution weights of each point along the local surface, which is more robust to various input data and greatly improves the performance of surface-style convolution.




As a local feature learning module, FPConv can be further integrated with other operations in classical nerual network architectures and works on various analysis tasks. We demonstrate FPConv on 3D object classification as well as 3D scene semantic segmentation. Networks with FPConv outperform previous surface-style approaches \cite{tatarchenko2018tangent}\cite{huang2019texturenet}\cite{pan2018convolutional} and achives comparable results with current start-of-the-art methods.  Moreover, our experiments also shows that FPConv performs better at regions that are relatively flat thus can be a complementary to volumetric-type works and joint training helps to boost the overall performance into state-of-the-art results. 



To summarize, the main contributions of this work are as follows:
\begin{itemize}
\item FPConv, a novel surface-style convolution for efficient 3D point cloud analysis.

\item Significant improvements over previous surface-style convolution based methods and comparable performance with state-of-the-art volumetric-style methods in classification and segmentation tasks.

\item An in-depth analysis and comparison between surface-style and volumetric-style convolution, demonstrating that they are complementary to each other and joint training boosts the performance into state-of-the-art.
\end{itemize}

\section{Related Work}



Deep learning based 3D data analysis has been a quite hot research topic in recent years. In this section, we mainly focus on point cloud analysis and briefly review previous works according to their underling methodologies.

\textbf{Volumetric-style point convolution} Since a point cloud disorderly distributes in a 3D space without any regular structures, pioneer works sample points into grids for conventional 3D convolutions apply, but limited by high computational load and low representation efficiency \cite{maturana2015voxnet, wu20153d, riegler2017octnet, song2017semantic}. PointNet \cite{qi2017pointnet} proposes a shared MLP on every point individually followed by a global max-pooling to extract global feature of the input point cloud. \cite{qi2017pointnet++} extends it with nested partitionings of point set to hierarchically learn more local features, and many works follow that to approximate point convolutions by MLPs \cite{li2018so, li2018pointcnn, hermosilla2018monte, wang2018deep}. However, adopting such a representation can not capture the local features very well. Recent works define explicit convolution kernels for points, whose weights are directly learned like image convolutions \cite{hua2018pointwise, xu2018spidercnn, groh2018flex, atzmon2018point, thomas2019kpconv}. Among them, KPConv \cite{thomas2019kpconv} proposes a spatially deformable point convolution with any number of kernel points which alleviates both varying densities and computational cost, outperform all associated methods on point analysis tasks. However, there volumetric-style approaches may not capture uniform areas very well.


\textbf{Graph-style point convolution} When the relationships among points have been established, a Graph-style convolution can be applied to explore and study point cloud more efficiently than volumetric-style. Convolution on a graph can be defined as convolution in its spectral domain.
\cite{bruna2013spectral, henaff2015deep, defferrard2016convolutional}. ChebNet \cite{defferrard2016convolutional} adopts Chebyshev polynomial basis for representing the spectral filters to alleviate the cost of explicitly computing the graph Fourier transform. Furthermore, \cite{kipf2016semi} uses a localized first-order approximation of spectral graph convolutions for semi-supervised learning on graph-structured data which greatly accelerates calculation efficiency and improves classification results. However, these methods are all depending on a specific graph structure. Then \cite{yi2017syncspeccnn} introduces a spectral parameterization of dilated convolution kernels and a spectral transformer network, sharing information across related but different shape structures. In the meantime, \cite{masci2015geodesic, bronstein2017geometric, simonovsky2017dynamic, monti2017geometric} focus on graph learning on manifold surface representation to avoid the spectral domain operation, while \cite{verma2018feastnet, wang2019dynamic} learn filters on edge relationships instead of points relative positions. Although a graph convolution combines features on local surface patches and can be invariant to the deformations in Euclidean space. However, reasonable relationships among distinct points are not easy to establish.

\textbf{Surface-style point convolution} Since data captured by 3D sensors typically represent surfaces, another mainstream approach attempts to operate directly on surface geometry. Most works project a shape surface consist of points to an intermediate grid structure, e.g. multi-view RGB-D images, following with conventional convolutions \cite{gupta2015indoor, li2016lstm, mccormac2017semanticfusion, boulch2017unstructured, lawin2017deep}. Such methods often suffer from the redundant representation of multi-view and the amubiguity casued of by different viewpoints. \cite{tatarchenko2018tangent} proposes projecting local neighborhoods of each point to its local tangent plane and processing them with 2D convolutions, which is efficient for analyzing dense point clouds of large-scale and outdoor environments. However, this method relies heavily on point tangent estimation, and this linear projection is not always optimal for complex areas. \cite{pan2018convolutional} optimizes the calculation with parallel tangential frames, while \cite{huang2019texturenet} utilizes a 4-rotational symmetric field to define a domain for convolution on surface, which not only increase the robustness, but also make the utmost of detailed information. However, existing surface-style learning algorithms cannot perform very well on challenge datasets such as S3DIS \cite{s3dis_cvpr16} and ScanNet \cite{dai2017scannet}, since they lose 1-dimensional information and they cannot estimate the surface accurately.

Our method is inspired by surface-style point convolutions. The network learns a non-linear projection for each local patch, say flattening the local neighborhood points into a 2D grid plane. Then 2D convolutions can be applied for feature extraction. Although learning on surface will lose 1-dimensional information, FPConv still achieves comparable performance with existing volumetric-style convolutions. In addition, our FPConv can be integrated into volumetric-style convolution and achieve state-of-the-art results.


\section{FPConv} \label{sec:3}

\begin{figure}[t]
\begin{center}
  \includegraphics[width=0.9\linewidth]{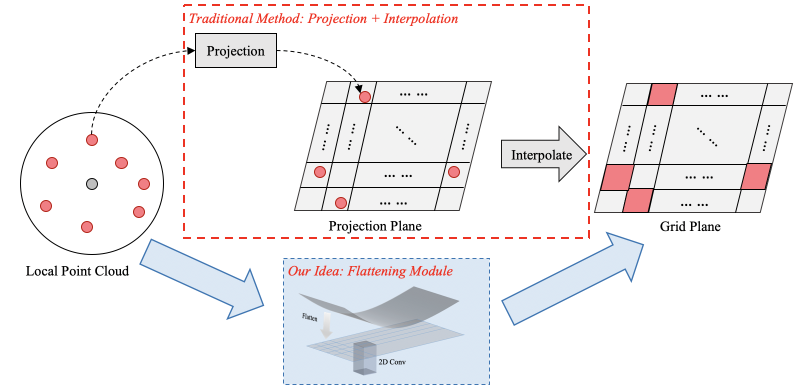}
\end{center}
\caption{Flattening module compared with traditional method: we design a module to learn local flattening directly instead of learning projection and interpolation separately.}
\label{fig:projection}
\end{figure}

\begin{figure*}[ht!]
\begin{center}
\includegraphics[width=0.9\linewidth]{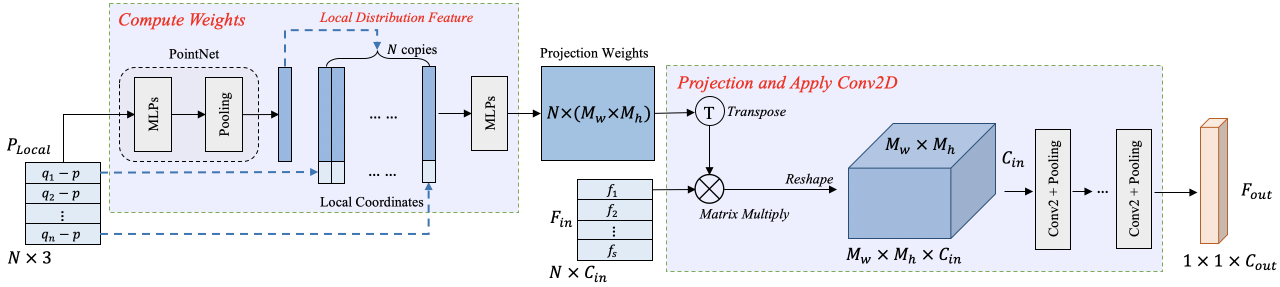}
\end{center}
  \caption{Process of conducting FPConv on local region centered around point $p$. The input coordinates and features come from $N$ neighbor points randomly picked in a radius range of $p$. The output is $F_{out}$ at $p$.}
\label{fig:fpconv}
\end{figure*}

In this section, we formally introduce FPConv. We first revisit the definition of convolution along point cloud surface and then show it can be simplified into a weight learning problem under discrete setting. All derivations are provided in the form of point clouds. 

\subsection{Learn Local Flattening}

\paragraph{Notation:} Let $p$ be  a point from a point cloud $\mathcal{P}$ and $\mathcal{F}(p)$ be a scalar function defined over points. Here $\mathcal{F}(p)$ can encode signals like color, geometry or  features from intermediate network layers. We denote  $\mathcal{N}(p)$ as a local point cloud patch centered at $p$ where $\mathcal{N}(p) = \Big\{q_i = q_i - p\ \Big\vert \ \ ||q_i-p||_2 < \rho , q_i \in \mathcal{P} \Big\}$ with $\rho \in \mathbb{R}$ being the chosen radius. 

\paragraph{Convolution on local surface:} In order to convolve $\mathcal{F}$ around the surface, we first extend it to a continuous function over a continuous surface. We introduce a virtual 2D plane $\mathcal{S}$ with a continuous signal $\mathcal{S}(u)$ together with a map $\pi(\cdot)$ which maps $\mathcal{N}(p)$ onto $\mathcal{S}$ and  

\begin{equation}
   \mathcal{S}(\pi(q_i))  = \mathcal{F}(q_i)
\label{eq:projection_0}
\end{equation}

The convolution at $p$ is defined as:

\begin{equation}
   \mathcal{X}(p)  = \int_\mathcal{S} c(u)\mathcal{S}(u)du 
\label{eq:projection_1}
\end{equation}

where $c(u)$ is a convolution kernel. We now describe how to formulate the above  convolution into a weight learning problem.

\paragraph{Local flattening by learning projection weights:} As shown in Eq.\ref{eq:projection_3}, with $\pi(\cdot)$,  $\mathcal{N}(p)$  will be mapped as scattered points in $\mathcal{S}$, thus we need a interpolation method to estimate the full signal function $S(u)$, as shown in Eq.\ref{eq:projection_3}.

\begin{equation}
   \mathcal{S}(u)  = \sum_{i}w\big(u, \pi(q_i)\big) \mathcal{S}\big(\pi(q_i)\big)
\label{eq:projection_3}
\end{equation}

Now if we  discretize $\mathcal{S}$ into a grid plane with size of $M_w \times M_h$. For each grid $\mathcal{S}(v_j)$, where $j$ in $\big\{1,2,...,M_w\times M_h\big\}$, we can have from Eq.\ref{eq:projection_0} and Eq.\ref{eq:projection_3}:

\begin{equation}
   \mathcal{S}(v_j)  = \sum_{i} w_{ji} \mathcal{F}(q_i)
\label{eq:projection_4}
\end{equation}

Where $w_{ji} = w\big(v_j, \pi(q_i)\big)$. Furthermore, we can rewrite Eq.\ref{eq:projection_1} in an approximate discretized form as:

\begin{equation}
\begin{split}
    \mathcal{X}(p)  &= \int_\mathcal{S} c(u)S(u)du  \\
                    &= \sum_j c_j \sum_{i} w_{ji} \mathcal{F}(q_i)  \\
                    &= M_{c} * \Big(W_f^T \times F(p)\Big)
\end{split}
\label{eq:projection_5}
\end{equation}

\begin{figure}[h!]
\begin{subfigure}{.23\textwidth}
		\captionsetup{labelformat=empty}
		\subcaption{Binary Sparsity}
		\centering
		\includegraphics[width=.7\linewidth]{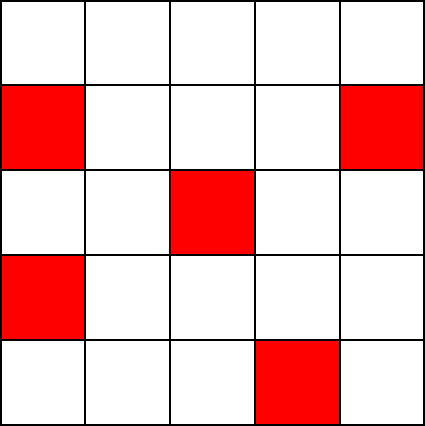}
\end{subfigure}
\begin{subfigure}{.23\textwidth}
		\captionsetup{labelformat=empty}
		\subcaption{Continuous Sparsity}
		\centering
		\includegraphics[width=.7\linewidth]{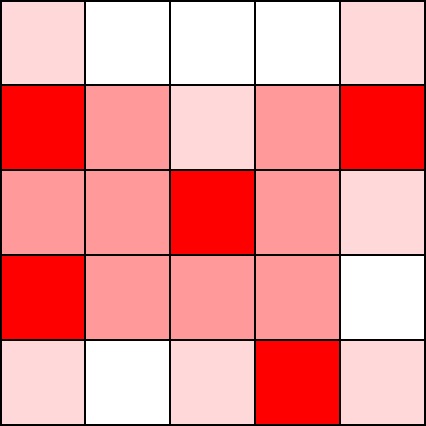}
\end{subfigure}%
\caption{Left: binary sparsity, intensity at each position should be 0 or 1. Right: continuous sparsity, intensity can be in the range of 0 to 1.}
\label{fig:sparsity}
\end{figure}

Where $c_j$ is discretized convolution kernel weights, and $j$ in $\big\{1,2,...,M_w\times M_h\big\}$.  Let $L = M_w\times M_h$, $W_f \in \mathbb{R}^{N\times L}$, $W_f (i, j) =  w\big(v_j, \pi(q_i)\big)$, and $F(p) = \Big(\mathcal{F}(q_1), ..., \mathcal{F}(q_N)\Big)^T \in \mathbb{R}^{N\times C}$. Now we can see that projection and interpolation can be combined into a single weight matrix $W_f$ where it only depends on the point location w.r.t the center point. 

\begin{figure*}[t]
\begin{center}
\includegraphics[width=0.75\linewidth]{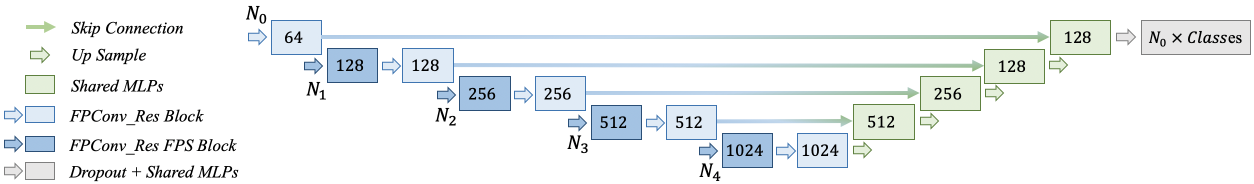}
\end{center}
  \caption{Network Architectures for Large Scene Segmentation: our segmentation architecture is composed of 4 downsampling layers for multi-scale analysis and apply skip connections for combination of features from encoder and decoder.}
\label{fig:architecture}
\end{figure*}

\subsection{Implementation} \label{sec:norm}

According to Eq.\ref{eq:projection_5}, we can design a module to learn projection weights directly instead of learning projection and interpolation separately, as shown in Fig.\ref{fig:projection}. We also want this module to have two properties: first, it should be invariant to input permutation since the local point cloud is unordered; second, it should be adaptive to input geometry, hence the projection should combine local coordinates and global information of local patch. Therefore, we first use pointnet \cite{qi2017pointnet} to extract the global feature of local region, namely distribution feature, which is invariant to permutation. Then we concatenate the distribution feature to each of the input points, as shown in Fig.\ref{fig:fpconv}. After that, a shared MLPs is employed to predict the final projection weights.


After projection, 2D convolution is applied on obtained grid feature plane.  To extract a local feature vector, global convolution or pooling can be applied on the last layer of 2D convolution network.




However, feature intensity of pixels in grid plane may be unbalanced when the summation of feature intensities received from points in local region is varying, which can break the stability of a neural network and make the training hard to converge.  In order to balance the feature intensity of grid plane, we further introduce two normalization methods on learned projection weights.


\textbf{Dense Grid Plane}: Let projection weights matrix be $W \in \mathbb{R}^{(N\times L)}$. One possible way to obtain a dense grid plane is normalizing $W$ at the first dimension by dividing their summation to make sure the summation of intensities received at each pixel is equal to 1. This is similar to bilinear interpolation method. In our implementation, we use softmax to avoid being divided by zero, which is shown in Eq.\ref{eq:softmax_norm}.

\begin{equation}
    W_{ij} = \frac{e^{W_{ij}}}{\sum_{k=1}^{N} e^{W_{kj}}}
\label{eq:softmax_norm}
\end{equation}

\textbf{Sparse Grid Plane}: 
Due to natural sparsity of point cloud, normalize the projection weights to get a dense grid plane may not be optimal. In this case, we design a 2-step normalization which can preserve the sparsity of projection weights matrix, and then the grid plane. Moreover, we conduct ablation studies on our proposed two normalization techniques.

\textit{First step} is to normalize at second dimension to balance the intensity given out by local neighbor points. Here, we add a positive $\epsilon$ to avoid being divided by zero. As shown in Eq.\ref{eq:noem_step1}, $W_{i \cdot}$ indicates $i$-th row of $W$.

\begin{equation}
    W_{ij} = \frac{W_{ij}}{||W_{i \cdot}||_2 + \epsilon}
\label{eq:noem_step1}
\end{equation}

\textit{Second step} is to normalize at first dimension to balance the intensity received at each pixel position. It can be implemented similar to first step by dividing by summation of each column. However, we choose another method shown in Eq.\ref{eq:noem_step2} to maintain a continuous sparsity, where $W_{\cdot j}$ indicates $j$-th column of $W$. Examples of continuous sparsity and binary sparsity are shown in Fig.\ref{fig:sparsity}.


\begin{equation}
    W_{ij} = \frac{W_{ij}}{\text{MAX}\Big(||W_{\cdot j}||_2,\ 1\Big)}
\label{eq:noem_step2}
\end{equation}

\section{Architecture}

\subsection{Residual FPConv Block}

\begin{figure}[t]
\begin{center}
   \includegraphics[width=0.95\linewidth]{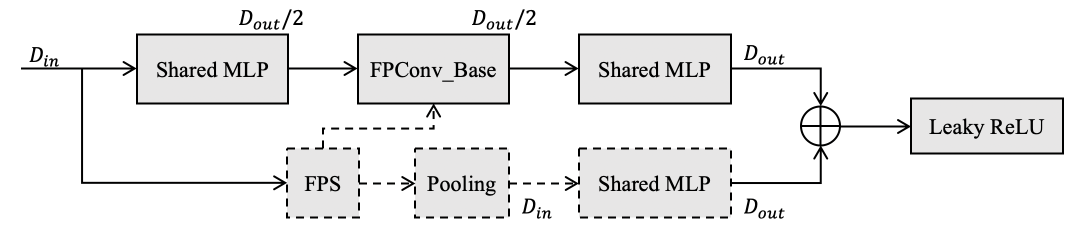}
\end{center}
   \caption{Residual FPConv Block: operations at the shortcut connection are optional, shared MLP is only required when $D_{out}$ is not equal to $D_{in}$, which is similar to projection shortcut \cite{He2015resnet}. FPS (Farthest Point Sampling \cite{qi2017pointnet++}) and Pooling is needed for downsampling.}
\label{fig:fpconv_res}
\end{figure}

To build a deep network for segmentation and classification, we develop a bottleneck-design residual FPConv block inspired by \cite{He2015resnet}, as shown in Fig.\ref{fig:fpconv_res}. This block takes a point cloud as input, applying a stack of shared MLP, FPConv, and shared MLP, where shared MLPs are responsible for reducing and then increasing (or restoring) dimensions, similar to $1\times 1$ convolutions in residual convolution block \cite{He2015resnet}.

\subsection{Multi-Scale Analysis}

As shown in Fig.\ref{fig:fpconv_res} and Fig.\ref{fig:architecture}, we design other operations for multi-scale analysis:

\textbf{Farthest Point Sampling}: we use iterative farthest point sampling to downsample the point cloud. As mentioned in PointNet++ \cite{qi2017pointnet++}, FPS has better coverage of the entire point set given the same number of centroids compared with random sampling.


\textbf{Pooling}: we use max-pooling to group local features. Given an input point cloud $\mathcal{P}_n$ and a downsampled point cloud $\mathcal{P}_m$ with their corresponding features $\mathcal{F}_n$ and $\mathcal{F}_m$, we group neighbors for each point in $\mathcal{P}_m$ with radius of $r$ and apply pooling operator on features of grouped points, as shown in Eq.\ref{eq:pooling}, where $\mathcal{P}_{neb} = \Big\{x\ \Big|\ ||x - y_i||_2 < r, x \in \mathcal{P}_n\Big\}$ for any $y_i \in \mathcal{P}_m$.

\begin{equation}
\mathcal{F}_{out}(y_i) = \text{Pooling}\Big(\mathcal{F}(\mathcal{P}_{neb})\Big)
    \label{eq:pooling}
\end{equation}

\textbf{FPConv with FPS}: similar to pooling operation, this block applies FPConv on each point of downsampled point cloud and search neighbors over full point cloud, as shown in Eq.\ref{eq:fpconv_fps}.

\begin{equation}
\mathcal{F}_{out}(y_i) = \text{FPConv}\Big(\mathcal{F}(\mathcal{P}_{neb})\Big)
    \label{eq:fpconv_fps}
\end{equation}

\textbf{Upsampling}: we use $K$ nearest neighbors interpolation to upsample point cloud by euclidean distance. Given a point cloud $\mathcal{P}_m$ with features $\mathcal{F}_m$ and a target point cloud $\mathcal{P}_n$, we compute feature for each point in $\mathcal{P}_n$ by interpolating its $K$ neighbor points searched over $\mathcal{P}_m$.

In the upsampling phase, skip connection and a shared MLPs is used for fusing features from encoder and decoder. $K$ nearest neighbors upsampling and shared MLPs can be replaced by de-convolution, but it does not lead to a significant improvement as mentioned in \cite{thomas2019kpconv}, so we do not employ it in our experiments.

Architecture shown in Fig.\ref{fig:architecture} is designed for large scene segmentation, including four layers of downsampling and upsampling for multi-scale analysis. For classification task, we apply a global pooling on the last layer of downsampling to obtain global feature for representing full point cloud, and then use a fully connected network for classification.

\subsection{Fusing Two Convolutions} \label{sec:4.3}

\begin{figure}[t]
\begin{center}
   \includegraphics[width=0.95\linewidth]{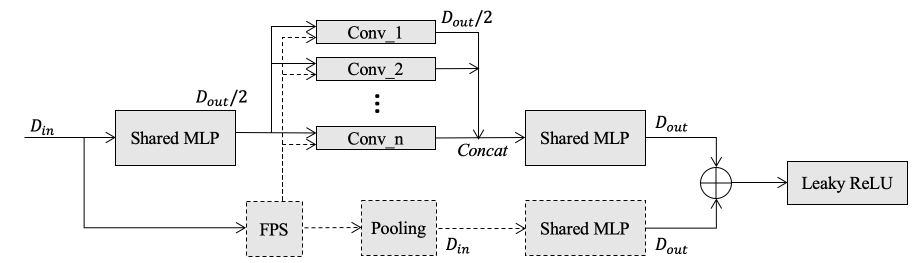}
\end{center}
   \caption{Parallel Residual Block: combine different types (Surface-Conv or Volume-Conv) of convolution kernel for fusion.}
\label{fig:parallel_res}
\end{figure}



As one of our main contributions, we also try to answer a question "Can we combine two convolutions for further boosting the performance?" The answer is yes but only works when the two convolutions are in different types or complementary (please see Section~\ref{sec:abl}), say surface-style and volumetric-style.  

In this section, we propose two convenient and quick fusion strategies, by combining two convolution operators in a single framework. First one is fusing different convolutional features, similar to inception net \cite{szegedy2015inception}. As shown in Fig.\ref{fig:parallel_res}, we design a parallel residual block. Given an input feature, apply multiple convolutions in parallel and then concatenate their outputs as fused feature. This strategy is suitable for some compatible methods, such like SA Module of PointNet++ \cite{qi2017pointnet++}, PointConv \cite{wu2019pointconv}, both using point cloud as input and applying downsampling strategy, which is the same used in our architecture.  

While for other incompatible methods, such as TextureNet \cite{huang2019texturenet} using mesh as an additional input, and KPConv \cite{thomas2019kpconv} applying grid downsampling, we have second fusion strategy by concatenating their output features in the last second layer of networks, an then applying a tiny network for fusion.

\begin{table}
	\setlength{\tabcolsep}{0.1em}
	\centering
	\begin{tabular}{l | c | c |ccc}
		\toprule
		Method           & Conv. & Samp. & ScanNet & S3DIS & S3DIS-6\\
		\midrule
		SPGraph \cite{landrieu2018large} & G & - & - & 58.0 & 62.1 \\
		ResGCN \cite{li2019deepgcns} & G & - & - & 53.8 & 60.0 \\
		HPEIN  \cite{jiang2019hierarchical}          & G & FPS & \textbf{61.8} & \textbf{61.9}  & \textbf{67.8}   \\
		\midrule
		PointNet \cite{qi2017pointnet}         & V & FPS & - & 41.1  & 47.6   \\
		PointNet++ \cite{qi2017pointnet++}      & V & FPS & 33.9    & -     & -      \\
		PointCNN \cite{li2018pointcnn}        & V & FPS & 45.8    & 57.3  & -      \\
		PointConv \cite{wu2019pointconv}      & V & FPS & 55.6 & 58.3$^\dag$ & - \\
		KPConv  \cite{thomas2019kpconv}   & V & Grid & \textbf{68.4}    & \textbf{65.4}  & \textbf{69.6}   \\
		\midrule
		TangentConv \cite{tatarchenko2018tangent}    & S & Grid & 43.8    & 52.6  & -      \\
		SurfaceConv \cite{pan2018convolutional}   & S & - & 44.2   & -     & -      \\
		TextureNet \cite{huang2019texturenet}      & S & QF \cite{huang2019texturenet} & 56.6    & -     & -      \\
		\textit{FPConv} (ours)           & S & FPS & \textbf{63.9}    & \textbf{62.8}  & \textbf{68.7}     \\
		\midrule
		FP $\oplus$ PointConv & S + V & - & - & 64.4 & - \\
		FP $\otimes$ PointConv & S + V & FPS & - & 64.8  & -      \\
		FP $\oplus$ KPConv & S + V & - & - & \textbf{66.7}  & -      \\
		
		\bottomrule
	\end{tabular}
	\caption{Mean IoU of large scene segmentation result. The second column is the convolution type (graph, surface or volumetric-style) and third column indicates sampling strategy. S3DIS-6 represents 6-fold cross validation. $\oplus$ is fusion in final feature level while $\otimes$ is fusion in convolutional feature level by applying parallel block. $^\dag$ indicates our implementation.}
	\label{table:segmentation}
\end{table}

\begin{table}
	\setlength{\tabcolsep}{0.25cm}
	\centering
	\begin{tabular}{l | c | c |ccc}
		\toprule
		Method           & Conv. & Samp. & Accuracy \\
		\midrule
		PointNet \cite{qi2017pointnet}         & V & FPS & 89.2   \\
		PointNet++ \cite{qi2017pointnet++}      & V & FPS & 90.7      \\
		PointCNN \cite{li2018pointcnn}        & V & FPS & 92.2      \\
		PointConv \cite{wu2019pointconv}      & V & FPS & 92.5 \\
		KPConv  \cite{thomas2019kpconv}   & V & Grid & 92.9  \\
		\hline
		\textit{FPConv} (ours)           & S & FPS & 92.5\\
		\bottomrule
	\end{tabular}
	\caption{Classification Accuracy on ModelNet40}
	\label{table:classification}
\end{table}

\begin{table*}[!ht]
	\setlength{\tabcolsep}{0.3em}
	\centering
	\begin{tabular}{l | ccc | ccccccccccccc}
		\toprule
		Method  & oA & mAcc & mIoU & ceil. & floor & wall & beam & col. & wind. & door & table & chair & sofa & book. & board & clut.\\
		\hline\hline
		PointConv$^\dag$ \cite{wu2019pointconv} & 85.4 & 64.7 & 58.3 & 92.8	& 96.3 & 77.0 & 0.0 & 18.2 & 47.7 & 54.3 & \textbf{87.9} & 72.8 & 61.6 & 65.9 & 33.9 & 49.3 \\
		KPConv \cite{thomas2019kpconv} & - & 70.9 & 65.4 & 92.6 & 97.3 & 81.4 & 0.0 & 16.5 & 54.4 & 69.5 & 80.2 & 90.1 & 66.4 & \textbf{74.6} & 63.7 & 58.1 \\
		\textit{FPConv} (ours) & 88.3 & 68.9 & 62.8 & 94.6 & 98.5 & 80.9 & 0.0 & 19.1 & 60.1 & 48.9 & 80.6 & 88.0 & 53.2 & 68.4 & 68.2 & 54.9 \\
		
		\hline
		FP $\otimes$ PointConv & 88.2 & 70.2 & 64.8 & 92.8 & 98.4 & 81.6 & 0.0 & 24.2 & 59.1 & 63.0 & 79.5 & 88.6	& \textbf{68.1} & 67.9 & 67.2 & 52.4 \\
		FP $\oplus$ PointConv & 88.6 & 71.5 & 64.4 & 94.2 & 98.5 & 82.4 & 0.0 & \textbf{25.5} & \textbf{62.9} & 63.1 & 79.8 & 87.9	& 53.5 & 68.3 & 67.1 & 54.5 \\
		KP $\oplus$ PointConv & 89.4 & 71.5 & 65.5 & \textbf{94.6} & 98.4 & 81.4 & 0.0 & 17.8 & 56.0 & \textbf{71.7} & 78.9 & \textbf{90.1} & 66.8 & 72.6 & 65.0 & \textbf{58.7} \\
		FP $\oplus$ KPConv & \textbf{89.9} & \textbf{72.8} & \textbf{66.7} & 94.5 & \textbf{98.6} & \textbf{83.9} & 0.0 & 24.5 & 61.1 & 70.9 & 81.6 & 89.4 & 60.3 & 73.5 & \textbf{70.8} & 57.8 \\
	
		\bottomrule
	\end{tabular}
	\caption{Detailed semantic segmentation scores on S3DIS Area-5. $\oplus$ represents fusion in final feature level while $\otimes$ represents fusion in convolutional feature level. Note that PointConv$^\dag$ indicates our implementation on S3DIS.}
	\label{table:s3dis_area5}
\end{table*}

\section{Experiments}

To demonstrate the efficacy of our proposed convolution, we conduct experiments on point cloud semantic segmentation and classification tasks. ModelNet40 \cite{wu20153d} is used for shape classification. Two large scale datasets named Stanford Large-Scale 3D Indoor Space (S3DIS)  \cite{s3dis_cvpr16} and ScanNet \cite{dai2017scannet} are used for 3D point cloud segmentation. We implement our FPConv with PyTorch \cite{paszke2017automatic}. Momentum gradient descent optimizer is used to optimize a point-wise cross entropy loss, with a momentum of 0.98, and an initial learning rate of 0.01 scheduled by cosine LR scheduler \cite{loshchilov2016sgdr}. Leaky ReLU and batch normalization are applied after each layer except the last fully connected layer. We trained our models 100 epochs for S3DIS, 300 epochs for ScanNet.







\subsection{3D Shape Classification}
ModelNet40 \cite{wu20153d} contains 12311 3D meshed models from 40 categories, with 9843 for training and 2468 for testing. Normal is used as additional input feature in our model. Moreover, randomly rotation among the $z$-axis and jittering are also used for data augmentation. As shown in Table.\ref{table:classification}, our model achieves state-of-the-art performance among surface-style methods.

\subsection{Large Scene Semantic Segmentation}
\label{sec:semseg}

\noindent
\textbf{Data}. S3DIS \cite{s3dis_cvpr16} contains 3D point clouds of 6 areas, totally 272 rooms. Each point in the scan is annotated with one of the semantic labels from 13 categories (chair, table, floor, wall etc. plus clutter). To prepare the training data, 14k points are randomly sampled from a randomly picked block of 2m by 2m. Both sampling are on-the-fly during training. While for testing, all points are covered. Each point is represented by a 9-dim vector of XYZ, RGB, and normalized location w.r.t to the room (from 0 to 1). In particular, the  sampling rate for each point is 0.5 in every training epoch.


ScanNet \cite{dai2017scannet} contains 1513 3D indoor scene scans, split into 1201 for training and 312 for testing. There are 21 classes in total and 20 classes are used for evaluation while 1 class for free space. Similar to S3DIS, we randomly sample the raw data in blocks then sample points on-the-fly during training. Each block is of size 2m by 2m, containing 11k points represented by a 6-dim vector, XYZ and RGB. 

\bigskip
\noindent
\textbf{Pipeline for fusion}. As mentioned in Section \ref{sec:4.3}, we propose two fusion strategies for fusing conv-kernels of different types. In our experiment, we select PointConv \cite{wu2019pointconv} and KPConv \cite{thomas2019kpconv} \textit{rigid} for comparison on S3DIS. We apply both two fusion strategies on PointConv with FPConv, and the second strategy, fusion on final feature level on FPConv with KPConv and PointConv with KPConv. In our experiments, KPConv \textit{rigid} is used for fusion, while its deformable version is ignored for missing released pre-trained model and hyper-parameters setting. Thus, in the latter part, we use KPConv to represent KPConv \textit{rigid}.

\begin{figure*}[ht]
	\begin{subfigure}{.2\textwidth}
		\captionsetup{labelformat=empty}
		\subcaption{Input}
		\centering
		\includegraphics[width=.95\linewidth]{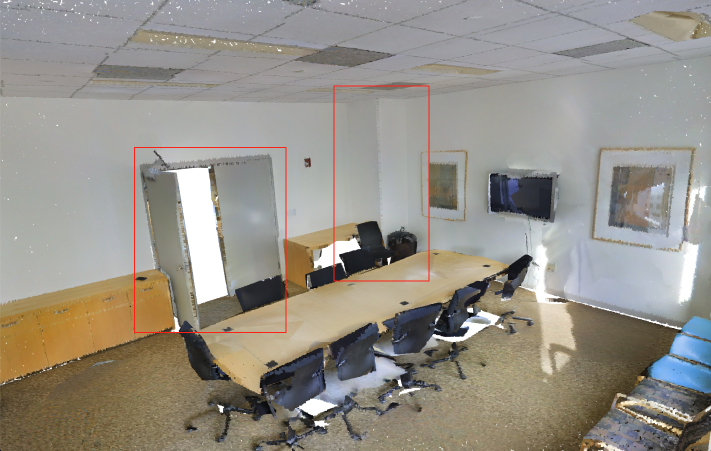}
	\end{subfigure}%
	\begin{subfigure}{.2\textwidth}
		\captionsetup{labelformat=empty}
		\subcaption{Ground Truth}
		\centering
		\includegraphics[width=.95\linewidth]{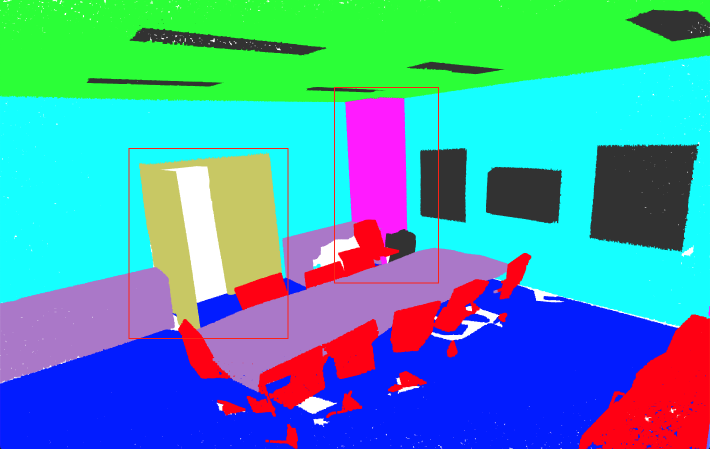}
	\end{subfigure}%
	\begin{subfigure}{.2\textwidth}
		\captionsetup{labelformat=empty}
		\subcaption{FP $\oplus$ KPConv}
		\centering
		\includegraphics[width=.95\linewidth]{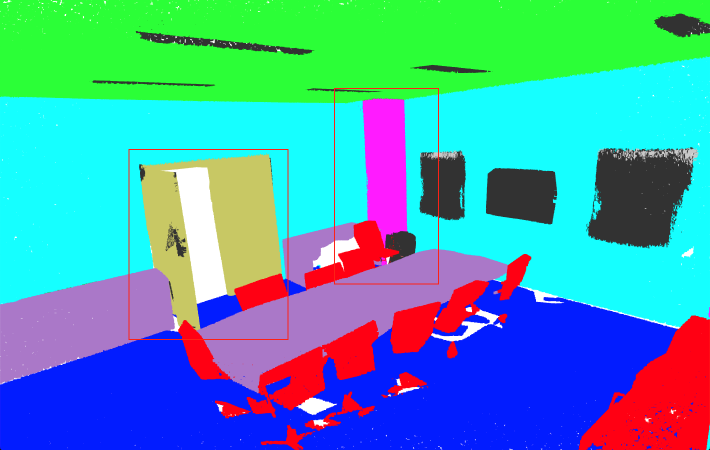}
	\end{subfigure}%
	\begin{subfigure}{.2\textwidth}
		\captionsetup{labelformat=empty}
		\subcaption{KPConv \cite{thomas2019kpconv}}
		\centering
		\includegraphics[width=.95\linewidth]{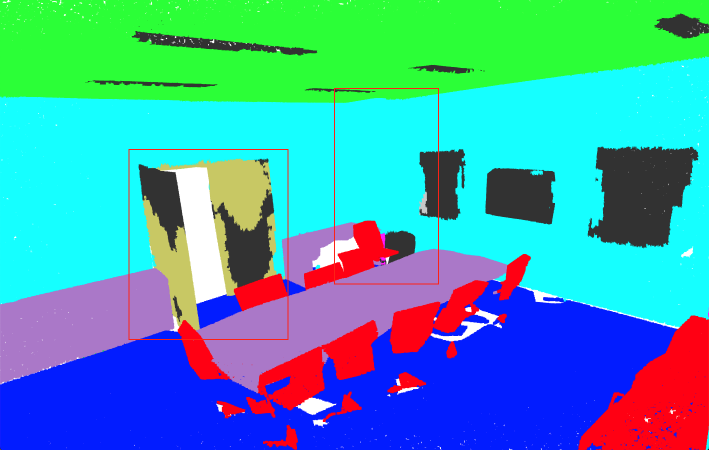}
	\end{subfigure}%
	\begin{subfigure}{.2\textwidth}
		\captionsetup{labelformat=empty}
		\subcaption{FPConv}
		\centering
		\includegraphics[width=.95\linewidth]{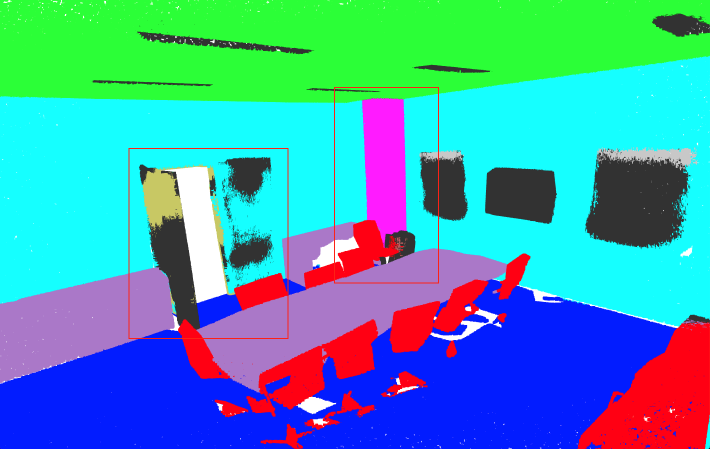}
	\end{subfigure}%
	
	\begin{subfigure}{.2\textwidth}
		\centering
		\includegraphics[width=.95\linewidth]{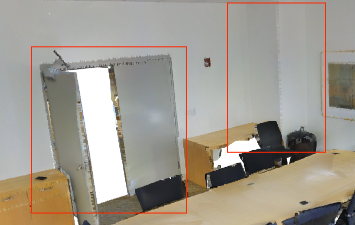}
	\end{subfigure}%
	\begin{subfigure}{.2\textwidth}
		\centering
		\includegraphics[width=.95\linewidth]{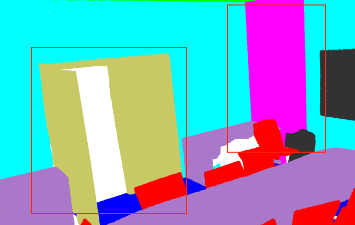}
	\end{subfigure}%
	\begin{subfigure}{.2\textwidth}
		\centering
		\includegraphics[width=.95\linewidth]{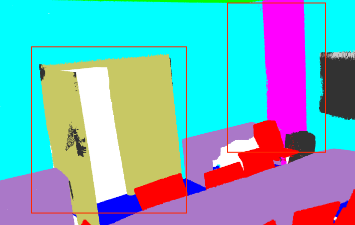}
	\end{subfigure}%
	\begin{subfigure}{.2\textwidth}
		\centering
		\includegraphics[width=.95\linewidth]{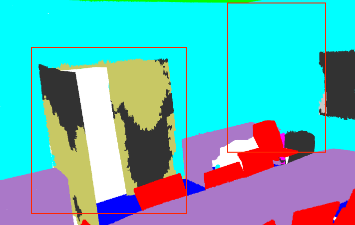}
	\end{subfigure}%
	\begin{subfigure}{.2\textwidth}
		\centering
		\includegraphics[width=.95\linewidth]{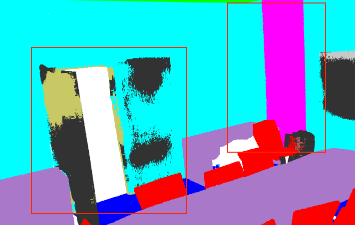}
	\end{subfigure}%
	
	\begin{subfigure}{.99\textwidth}
		\centering
		\includegraphics[width=.9\linewidth]{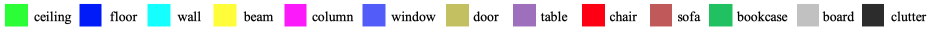}
	\end{subfigure}%
	\caption{Visualization of semantic segmentation results of S3DIS area 5. Images shown in second row is roomed version of first row images. The two red bounding boxes show that two structures that KPConv \cite{thomas2019kpconv} and FPConv cannot handle both of them very well while FP $\oplus$ KPConv can do it much better.}
	\label{fig:s3dis}
\end{figure*}


\begin{figure}[t]
\begin{subfigure}{.15\textwidth}
		\captionsetup{labelformat=empty}
		\subcaption{Input}
		\centering
		\includegraphics[width=.98\linewidth]{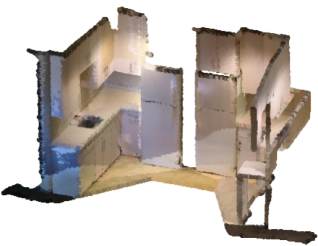}
\end{subfigure}
\begin{subfigure}{.15\textwidth}
		\captionsetup{labelformat=empty}
		\subcaption{Ground Truth}
		\centering
		\includegraphics[width=.98\linewidth]{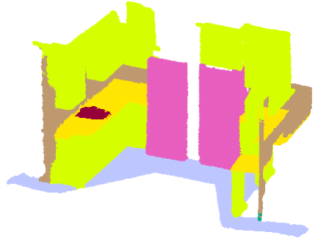}
\end{subfigure}
\begin{subfigure}{.15\textwidth}
		\captionsetup{labelformat=empty}
		\subcaption{Prediction}
		\centering
		\includegraphics[width=.98\linewidth]{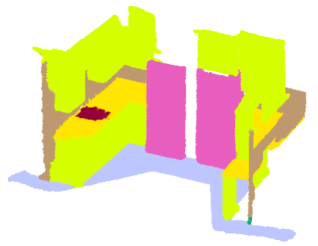}
\end{subfigure}%

\begin{subfigure}{.15\textwidth}
		\captionsetup{labelformat=empty}
		\centering
		\includegraphics[width=.98\linewidth]{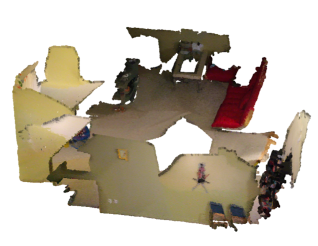}
\end{subfigure}
\begin{subfigure}{.15\textwidth}
		\captionsetup{labelformat=empty}
		\centering
		\includegraphics[width=.98\linewidth]{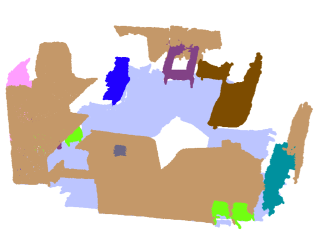}
\end{subfigure}
\begin{subfigure}{.15\textwidth}
		\captionsetup{labelformat=empty}
		\centering
		\includegraphics[width=.98\linewidth]{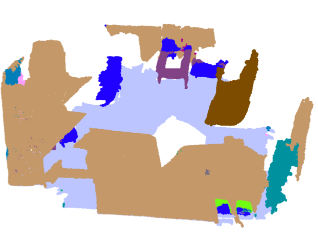}
\end{subfigure}%
\caption{Visualization of segmentation results on ScanNet.}
\label{fig:vis_scannet}
\end{figure}

\bigskip
\noindent
\textbf{Results}. Following \cite{qi2017pointnet}, we report the results on two settings for S3DIS, the first one is evaluation on Area 5, and another one is 6-fold cross validation (calculating the metrics with results from different folds merged). We report the mean of class-wise intersection over union (mIoU), overall point-wise accuracy (oA) and the mean of class-wise accuracy (mAcc). For Scannet \cite{dai2017scannet}, we report the mIoU score tested on ScanNet bencemark.

Results (mIoU) are shown in Table.\ref{table:segmentation}. Detailed results of S3DIS including mIoU of each class are shown in Table.\ref{table:s3dis_area5}. As we can see, FPConv outperforms all the existing surface-style learning methods with large margins. Specifically, the mIoU of FPConv on Scannet \cite{dai2017scannet} benchmark reaches 63.9\%, which outperforms the previous best surface-style method by 7.3\%. In addition, our FPConv fused with KPConv achieves state-of-the-art performance on S3DIS. 

Even though mIoU of S3DIS of our FPConv is lower than KPConv, there are still IoUs of some classes outperform the ones of KPConv, such as ceiling, floor, board, etc. Particularly, we find that all of these classes are flat objects, which should have small curvatures. Based on this discovery, we further conduct several ablation studies to explore the relationship between segmentation performance of FPConv and objects curvatures, as shown in next section. Visualization of results are shown in Fig.\ref{fig:s3dis} for S3DIS and Fig.\ref{fig:vis_scannet} for ScanNet.

%


\section{Ablation Study}
\label{sec:abl}
Two ablation studies are conducted, the first one is exploring fusion of surface-style and volumetric-style convolutions. Another one is the effect of detailed configurations, normalization methods and plane size on FPConv.


\subsection{On Fusion of S.Conv and V.Conv}
We firstly study the performance for different combination methods of the two convolutions. Before that, we show an experimental finding that they are complementary and good at analyzing different specific scenes. 

\begin{figure}[t]
\begin{subfigure}{.23\textwidth}
		\captionsetup{labelformat=empty}
		\centering
		\includegraphics[width=.95\linewidth]{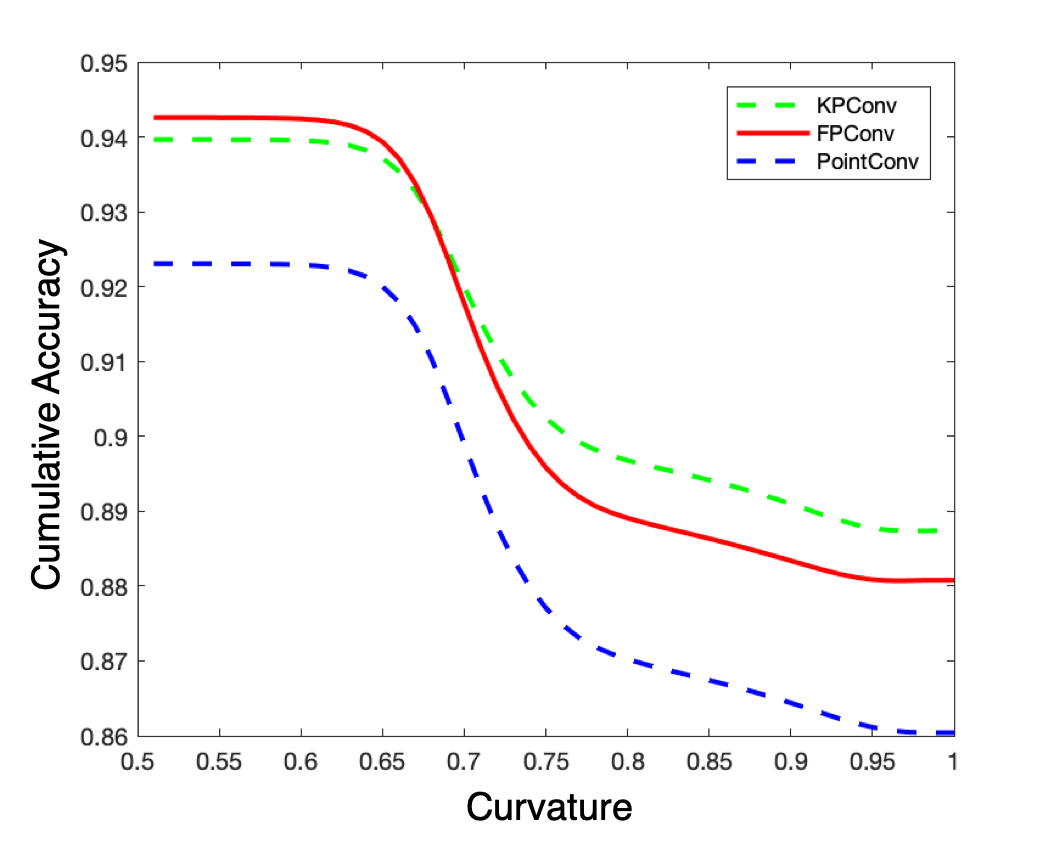}
\end{subfigure}
\begin{subfigure}{.23\textwidth}
		\captionsetup{labelformat=empty}
		\centering
		\includegraphics[width=.99\linewidth]{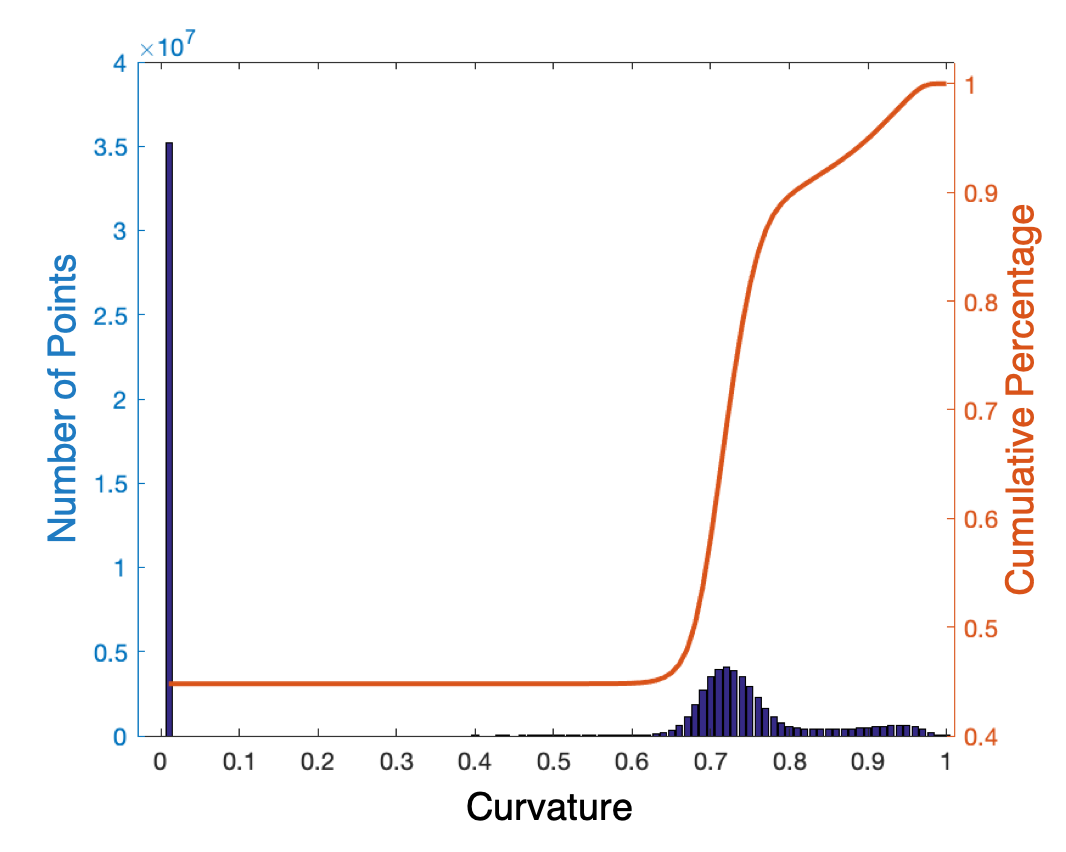}
\end{subfigure}%
\caption{Left: curvature versus cumulative accuracy. Right: histogram of curvatures.}
\label{fig:cur1}
\end{figure}

\begin{figure}[t]
\begin{center}
   \includegraphics[width=0.9\linewidth]{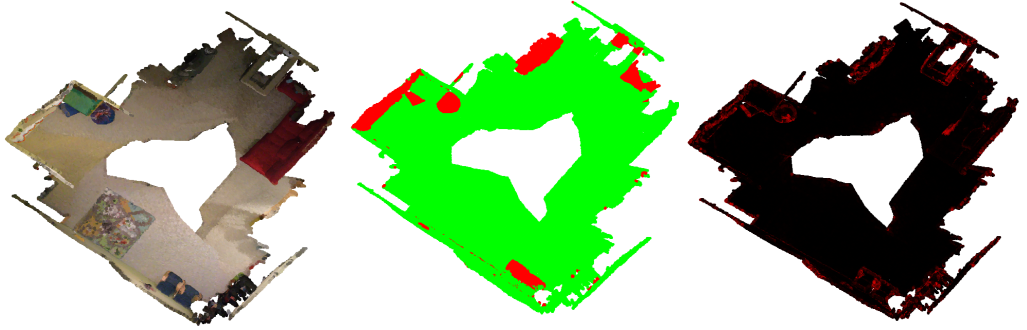}
\end{center}
\caption{Relationship between accuracy and curvature. Left: raw point cloud. Middle: prediction of FPConv with incorrect points highlighted in red, and correct in green. Right: points with large curvature are highlighted in red. We can see that distribution of incorrect points is consistent with large curvature points. We can see that distribution of incorrect points is consistent with large curvature points.}
\label{fig:cur_vis}
\end{figure}

\paragraph{Performance vs. Curvature} As experiments mentioned in Section \ref{sec:semseg}, we claim that FPConv can perform better on area with small curvature. To be more convincing, we analyzed the relationship between overall accuracy and curvatures, which is shown in the left of Fig.\ref{fig:cur1}. We can see that FPConv outperforms PointConv \cite{wu2019pointconv} and KPConv \cite{thomas2019kpconv} when curvatures are small, and FPConv cannot perform very well on structures which have large curvatures. Moreover, the histogram of distribution of points curvatures shown in the right of Fig.\ref{fig:cur1} implies almost all points have either large curvatures or small curvatures. This explains why there is a huge performance degradation when curvature increases. Furthermore, as shown in Fig.\ref{fig:cur_vis}, we highlight points (in red) with incorrect prediction, and points (in red) with large curvature. It is oblivious that incorrect prediction is concentrated on area with large curvature and FPConv performs well in flat area.




\paragraph{Ablation analysis on fusion method} As mentioned above, FPConv which is a surface-style convolution performs better in flat area, worse in rough area and KPConv, as a volumetric-style convolution performs oppositely. We believe that they can be complementary to each other and conduct 4 fusion experiments, FPConv $\otimes$ PointConv, FPConv $\oplus$ PointConv, KPConv $\oplus$ PointConv, and FPConv $\oplus$ KPConv, where $\oplus$ represents fusion in final feature level and $\otimes$ represents fusion in conv level. We don't conduct fusion of FPConv and KPConv in conv level for their incompatible downsampling strategies. As shown in Table.\ref{table:s3dis_area5}, fusion of FPConv with PointConv or KPConv brings a great improvement, while fusion of PointConv with KPConv brings little improvement. Therefore, we can claim that our FPConv can be complementary to volumetric-style convolutions, which may direct the convolution design for point cloud in the future.


Visualization of results are shown in Fig.\ref{fig:s3dis}. Our FPConv can capture better flat structures than KPConv, such as the class column that does not shown in KPConv. While KPConv can capture better complex structures, such as the door. Moreover, the fusion of KPConv and FPConv can achieve better results than both KPConv and FPConv.

\begin{table}
	\setlength{\tabcolsep}{0.33cm}
	\centering
	\begin{tabular}{l | c | c | c }
		\toprule
		Method            &  mIoU & mAcc & oAcc \\
		\midrule
		w sparse norm + 6x6 & \textbf{62.8} & \textbf{69.0} & 88.3 \\
		w dense norm + 6x6 & 61.6 & 68.5 & 87.6 \\
		w/o norm + 6x6 & 59.8 & 67.1 & 86.2 \\
		w sparse norm + 5x5 & 61.8 & 68.1 & \textbf{88.4} \\
		\bottomrule
	\end{tabular}
	\caption{Different normalization results on S3DIS area 5. 6x6 and 5x5 represent different plane sizes.}
	\label{table:normal}
\end{table}

\subsection{On FPConv Architecture Design}

We conduct 4 experiments as shown in Table.\ref{table:normal}, to study influence of normalization method and the size of grid plane on performance of FPConv. It tells us that, sparse-norm which indicates 2-step normalization method mentioned in Section \ref{sec:norm} performs better than dense-norm. In addition, higher resolution of grid plane may achieve better performance, while bring higher memory cost as well.

\section{Conclusion}

In this work, we propose FPConv, a novel surface-style convolution operator on 3D point cloud. FPConv takes a local region of point cloud as input, and flattens it onto a 2D grid plane by predicting projection weights, followed by regular 2D convolutions. Our experiments demonstrate that FPConv significantly improved the performance of surface-style convolution methods. Furthermore, we discover that surface-style convolution can be a complementary to volumetric-style convolution and jointly training can boost the performance into state-of-the-art. We believe that surface-style convolutions can play an important role in feature learning of 3D data and is a promising direction to explore. 



\paragraph{Acknowledge} This work was supported in part by gra-nts No.2018YFB1800800, 
No.ZDSYS201707251409055, NSFC-61902334, NSFC-61629101, No.2018B030338001, and No.2017ZT07X152.


{\small
\bibliographystyle{latex/ieee_fullname}
\bibliography{latex/egbib_sup}
}

 \section*{Supplementary}

\paragraph{The supplementary material contains:}

\begin{itemize}
	\item The results of the proposed fusion model between FPConv and PointConv \cite{wu2019pointconv} on Scannet \cite{dai2017scannet}.
	\item The results of the proposed fusion model between FPConv and KPConv-\textit{deform} \cite{thomas2019kpconv} on S3DIS \cite{s3dis_cvpr16}.
    \item More qualitative and quantitative results on large-scale scene segmentation tasks.
\end{itemize}

\section*{A. More results of the proposed fusion strategy}

\subsection*{a. Fusing FPConv and PointConv on ScanNet}

We conduct experiments on fusion of FPConv with PointConv \cite{wu2019pointconv} on ScanNet \cite{dai2017scannet}. The results are reported in Table.\ref{table:fuse_scannet}, where all methods are performed under same settings (architecture, hyper parameters, etc.). Note that we reduce sampled points to 8k in a block of 1.5m $\times$ 1.5m for all experiments.

\begin{table} [h!]
	\setlength{\tabcolsep}{0.35cm}
	\centering
	\begin{tabular}{l | c | c |ccc}
		\toprule
		Method                             & mIoU & mA   & oA \\
		\midrule
		\hline
		PointConv \cite{wu2019pointconv}   & 55.6 & -      & -   \\
		PointConv$^\dag$                   & 60.3 & 72.3 & 83.6  \\
		\textit{FPConv} (ours)             & 63.0 & 75.6 & 85.3  \\
		\hline
		FPConv $\otimes$ PointConv         & \textbf{64.2} & \textbf{76.1} & \textbf{86.0}  \\
		\bottomrule
	\end{tabular}
	\caption{Quantitative results of the segmentation task on evaluation dataset of ScanNet. PointConv$^\dag$ indicates our re-implementation of PointConv \cite{wu2019pointconv}.}
	\label{table:fuse_scannet}
	 
\end{table}

\subsection*{b. Fusing FPConv and  KPConv-\textit{deform} on S3DIS}


We further report the results of the proposed fusion model between FPConv and KPConv-\textit{deform} \cite{thomas2019kpconv} on S3DIS \cite{s3dis_cvpr16} in Table.\ref{table:sup_s3dis_fuse}, where the results of each class are also shown. As seen, the proposed fusion model wins all existing methods, reaching the state-of-the-art.

\section*{B. Parameter Comparison}

We compared the trainable parameters of PointConv, FPConv and their fusion forms in Table.\ref{table:param_comp}. We can see that fusion of convolution operators of same type cannot bring significant improvement and even get worse. While for the fusion of different types (FPConv $\otimes$ PointConv), even if we reduce the channel size of the fusion block, it still performs much better than before the fusion.

\begin{table} [h!]
	\setlength{\tabcolsep}{0.15cm}
	\centering
	\begin{tabular}{l | c | c cc}
		\toprule
		Method                             & mIoU & parameters\\
		\midrule
		\hline
		PointConv$^\dag$                   & 60.3 & 4.5    \\
		\textit{FPConv} (ours)             & 63.4 & 4.8    \\
		\hline
		FPConv $\otimes$ PointConv             & 64.2 & 7.7   \\
		FPConv $\otimes$ PointConv + mid ch / 2  & \textbf{65.1} & \textbf{3.8}   \\
		PointConv $\otimes$ PointConv & 60.7 & 7.6   \\
		FPConv $\otimes$ FPConv & 62.9 & 7.8   \\
		\bottomrule
	\end{tabular}
	\caption{Comparison of trainable parameters between different convolution operators on ScanNet evaluation dataset. $^\dag$ indicates our implementation. + mid ch /2 is halving the middle channel size of bottleneck in residual block.}
	\label{table:param_comp}
	 
\end{table}

\section*{C. More Results on Segmentation Tasks}

\begin{figure*} [t!]
    \centering
    \includegraphics[width=.95\linewidth]{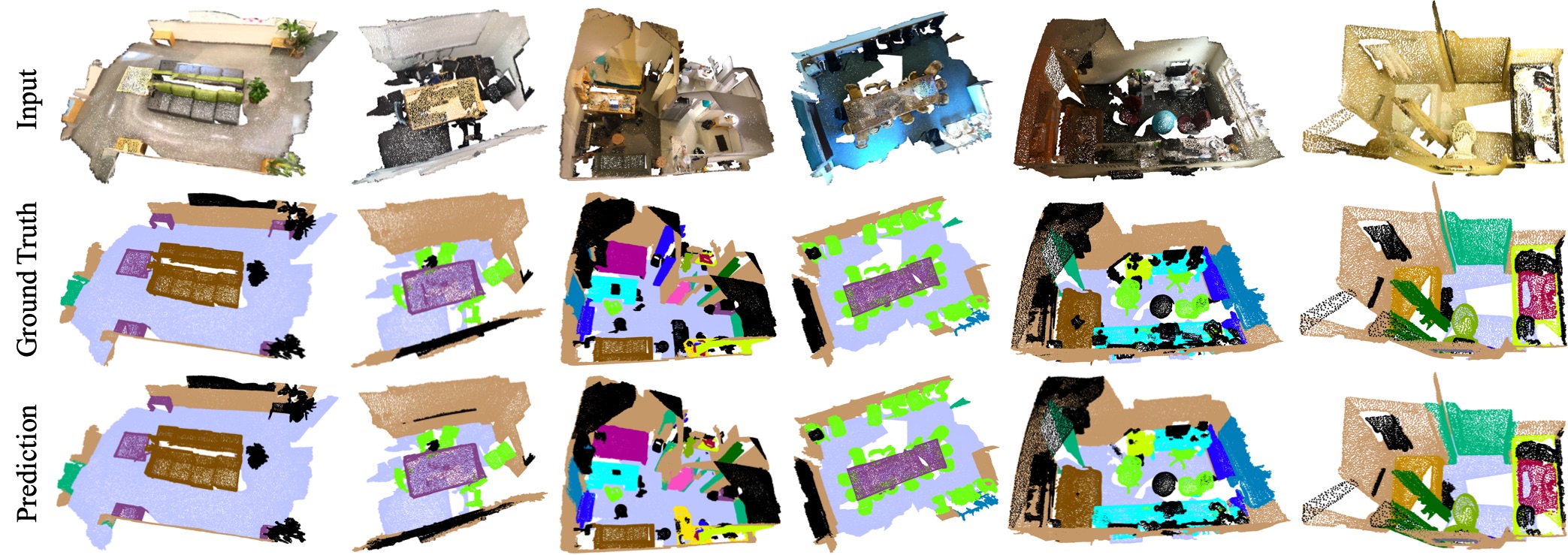}
    \caption{Visualization of semantic segmentation results of FPConv on ScanNet.}
    \label{fig:sup_vis_scannet}
\end{figure*}

\begin{figure*}[t!]
	\begin{subfigure}{.2\textwidth}
		\captionsetup{labelformat=empty}
		\subcaption{Input}
		\centering
		\includegraphics[width=.95\linewidth]{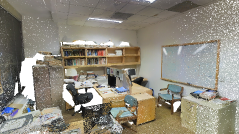}
	\end{subfigure}%
	\begin{subfigure}{.2\textwidth}
		\captionsetup{labelformat=empty}
		\subcaption{Ground Truth}
		\centering
		\includegraphics[width=.95\linewidth]{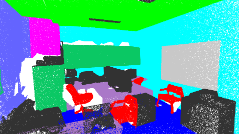}
	\end{subfigure}%
	\begin{subfigure}{.2\textwidth}
		\captionsetup{labelformat=empty}
		\subcaption{FPConv $\oplus$ KPConv}
		\centering
		\includegraphics[width=.95\linewidth]{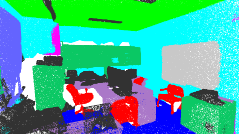}
	\end{subfigure}%
	\begin{subfigure}{.2\textwidth}
		\captionsetup{labelformat=empty}
		\subcaption{KPConv}
		\centering
		\includegraphics[width=.95\linewidth]{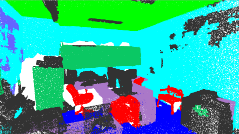}
	\end{subfigure}%
	\begin{subfigure}{.2\textwidth}
		\captionsetup{labelformat=empty}
		\subcaption{FPConv}
		\centering
		\includegraphics[width=.95\linewidth]{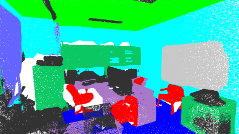}
	\end{subfigure}%
	
	\begin{subfigure}{.2\textwidth}
		\centering
		\includegraphics[width=.95\linewidth]{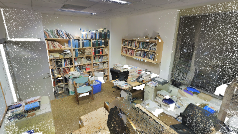}
	\end{subfigure}%
	\begin{subfigure}{.2\textwidth}
		\centering
		\includegraphics[width=.95\linewidth]{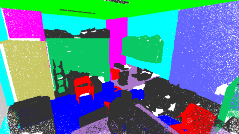}
	\end{subfigure}%
	\begin{subfigure}{.2\textwidth}
		\centering
		\includegraphics[width=.95\linewidth]{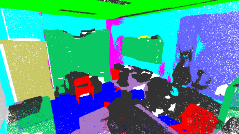}
	\end{subfigure}%
	\begin{subfigure}{.2\textwidth}
		\centering
		\includegraphics[width=.95\linewidth]{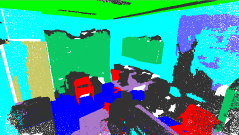}
	\end{subfigure}%
	\begin{subfigure}{.2\textwidth}
		\centering
		\includegraphics[width=.95\linewidth]{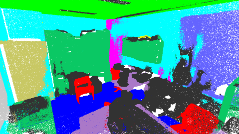}
	\end{subfigure}%
	
		\begin{subfigure}{.2\textwidth}
		\centering
		\includegraphics[width=.95\linewidth]{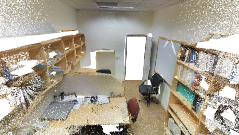}
	\end{subfigure}%
	\begin{subfigure}{.2\textwidth}
		\centering
		\includegraphics[width=.95\linewidth]{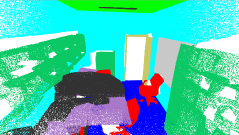}
	\end{subfigure}%
	\begin{subfigure}{.2\textwidth}
		\centering
		\includegraphics[width=.95\linewidth]{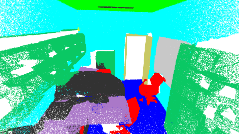}
	\end{subfigure}%
	\begin{subfigure}{.2\textwidth}
		\centering
		\includegraphics[width=.95\linewidth]{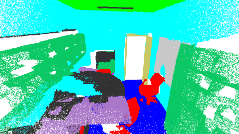}
	\end{subfigure}%
	\begin{subfigure}{.2\textwidth}
		\centering
		\includegraphics[width=.95\linewidth]{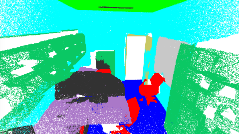}
	\end{subfigure}%

		\begin{subfigure}{.2\textwidth}
		\centering
		\includegraphics[width=.95\linewidth]{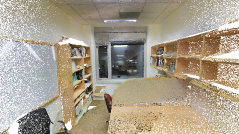}
	\end{subfigure}%
	\begin{subfigure}{.2\textwidth}
		\centering
		\includegraphics[width=.95\linewidth]{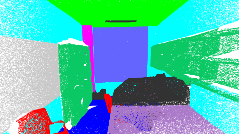}
	\end{subfigure}%
	\begin{subfigure}{.2\textwidth}
		\centering
		\includegraphics[width=.95\linewidth]{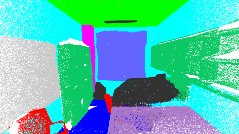}
	\end{subfigure}%
	\begin{subfigure}{.2\textwidth}
		\centering
		\includegraphics[width=.95\linewidth]{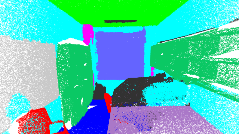}
	\end{subfigure}%
	\begin{subfigure}{.2\textwidth}
		\centering
		\includegraphics[width=.95\linewidth]{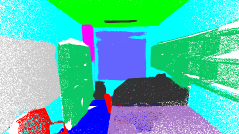}
	\end{subfigure}%

	\begin{subfigure}{.2\textwidth}
	\centering
	\includegraphics[width=.95\linewidth]{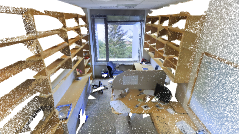}
	\end{subfigure}%
	\begin{subfigure}{.2\textwidth}
		\centering
		\includegraphics[width=.95\linewidth]{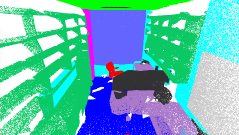}
	\end{subfigure}%
	\begin{subfigure}{.2\textwidth}
		\centering
		\includegraphics[width=.95\linewidth]{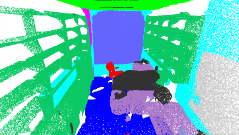}
	\end{subfigure}%
	\begin{subfigure}{.2\textwidth}
		\centering
		\includegraphics[width=.95\linewidth]{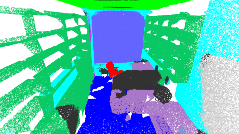}
	\end{subfigure}%
	\begin{subfigure}{.2\textwidth}
		\centering
		\includegraphics[width=.95\linewidth]{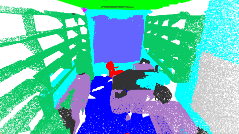}
	\end{subfigure}%
	
		\begin{subfigure}{.2\textwidth}
		\centering
		\includegraphics[width=.95\linewidth]{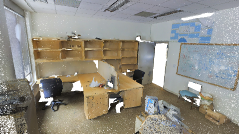}
	\end{subfigure}%
	\begin{subfigure}{.2\textwidth}
		\centering
		\includegraphics[width=.95\linewidth]{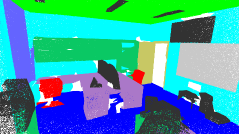}
	\end{subfigure}%
	\begin{subfigure}{.2\textwidth}
		\centering
		\includegraphics[width=.95\linewidth]{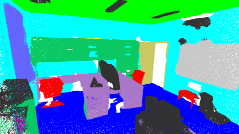}
	\end{subfigure}%
	\begin{subfigure}{.2\textwidth}
		\centering
		\includegraphics[width=.95\linewidth]{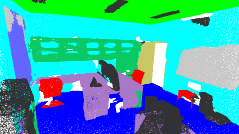}
	\end{subfigure}%
	\begin{subfigure}{.2\textwidth}
		\centering
		\includegraphics[width=.95\linewidth]{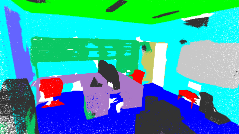}
	\end{subfigure}%
	
		\begin{subfigure}{.2\textwidth}
		\centering
		\includegraphics[width=.95\linewidth]{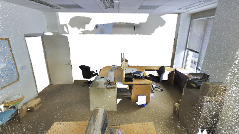}
	\end{subfigure}%
	\begin{subfigure}{.2\textwidth}
		\centering
		\includegraphics[width=.95\linewidth]{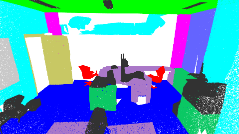}
	\end{subfigure}%
	\begin{subfigure}{.2\textwidth}
		\centering
		\includegraphics[width=.95\linewidth]{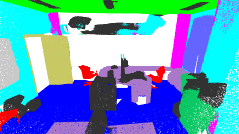}
	\end{subfigure}%
	\begin{subfigure}{.2\textwidth}
		\centering
		\includegraphics[width=.95\linewidth]{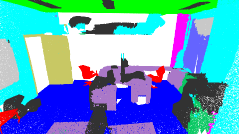}
	\end{subfigure}%
	\begin{subfigure}{.2\textwidth}
		\centering
		\includegraphics[width=.95\linewidth]{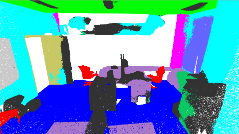}
	\end{subfigure}%
	
	\begin{subfigure}{.99\textwidth}
		\centering
		\includegraphics[width=.9\linewidth]{latex/images/color.png}
	\end{subfigure}%
	\caption{Qualitative comparisons of semantic segmentation tasks on S3DIS area 5. $\oplus$ indicates fusing in final feature level.}
	\label{fig:sdis_sup_vis}
\end{figure*}

We provide more details of our experimental results. As shown in Table.\ref{table:s3dis_6fold}, we compare our FPConv with other popular methods on S3DIS \cite{s3dis_cvpr16} 6-fold cross validation, which shows that FPConv can achieve higher score on flat-shaped objects, such like ceiling, floor, table, board, etc. While KPConv \cite{thomas2019kpconv}, a volumetric-style method, performs better on complex structures. More visual results are shown in Fig.\ref{fig:sup_vis_scannet} and Fig.\ref{fig:sdis_sup_vis}.

\begin{table*}[t!]
	\setlength{\tabcolsep}{0.35em}
	\centering
	\begin{tabular}{l | c | ccccccccccccc}
		\toprule
		Method  & mIoU & ceil. & floor & wall & beam & col. & wind. & door & table & chair & sofa & book. & board & clut.\\
		\hline\hline
        KPConv-\textit{rigid} \cite{thomas2019kpconv} & 65.4 & 92.6 & 97.3 & 81.4 & 0.0 & 16.5 & 54.4 & \textbf{69.5} & 80.2 & 90.1 & 66.4 & 74.6 & 63.7 & 58.1 \\
        KPConv-\textit{deform} \cite{thomas2019kpconv} &67.1 & 92.8 & 97.3 & \textbf{82.4} & 0.0 & \textbf{23.9} & 58.0 & 69.0 & \textbf{81.5} & \textbf{91.0} & \textbf{75.4} & \textbf{75.3} & 66.7 & \textbf{58.9} \\
		\textit{FPConv} (ours) & 62.8 & \textbf{94.6} & \textbf{98.5} & 80.9 & 0.0 & 19.1 & \textbf{60.1} & 48.9 & 80.6 & 88.0 & 53.2 & 68.4 & \textbf{68.2} & 54.9 \\
		\hline\hline
		FP $\oplus$ KP-\textit{rigid}& 66.7 & \textbf{94.5} & \textbf{98.6} & \textbf{83.9} & 0.0 & 24.5 & 61.1 & \textbf{70.9} & 81.6 & 89.4 & 60.3 & 73.5 & \textbf{70.8} & 57.8 \\
		FP $\oplus$ KP-\textit{deform}& \textbf{68.2} & 94.2 & 98.5 & 83.7 & 0.0 & \textbf{24.7} &\textbf{ 63.0} & 66.6 & \textbf{82.5} & \textbf{91.9} & \textbf{76.7} & \textbf{75.6} & 70.5 & \textbf{59.1} \\
		\bottomrule
	\end{tabular}
	\caption{Fusion results on S3DIS area 5. $\oplus$ indicates fusing in final feature level.}
	\label{table:sup_s3dis_fuse}
\end{table*}

\begin{table*}[t!]
	\setlength{\tabcolsep}{0.35em}
	\centering
	\begin{tabular}{l | c | ccccccccccccc}
		\toprule
		Method  & mIoU & ceil. & floor & wall & beam & col. & wind. & door & table & chair & sofa & book. & board & clut.\\
		\hline\hline
        PointNet \cite{qi2017pointnet} &47.6 &88.0 &88.7 &69.3 &42.4 &23.1 &47.5 &51.6 &54.1 &42.0 &9.6 &38.2 &29.4 &35.2 \\
        RSNet \cite{huang2018recurrent} &56.5 &92.5 &92.8 &78.6 &32.8 &34.4 &51.6 &68.1 &59.7 &60.1 &16.4 &50.2 &44.9 &52.0 \\ 
        SPGraph \cite{landrieu2018large} &62.1 &89.9 &95.1 &76.4 &62.8 &47.1 &55.3 &68.4 &69.2 &73.5 &45.9 &63.2 &8.7 &52.9 \\
        PointCNN \cite{li2018pointcnn} &65.4 &94.8 &97.3 &75.8 &63.3 &51.7 &58.4 &57.2 &69.1 &71.6 &61.2 &39.1 &52.2 &58.6 \\
        HPEIN \cite{jiang2019hierarchical} & 67.8 &-&-&-&-&-&-&-&-&-&-&-&-&-\\
        KPConv-\textit{rigid} \cite{thomas2019kpconv}  &69.6 &93.7 &92.0 &82.5 &62.5 &49.5 &65.7 &\textbf{77.3} &64.0 &57.8 &71.7 &68.8 &60.1 &59.6 \\
        KPConv-\textit{deform} \cite{thomas2019kpconv} &\textbf{70.6} &93.6 &92.4 &\textbf{83.1} &\textbf{63.9} &\textbf{54.3} &\textbf{66.1} &76.6 &64.0 &57.8 &\textbf{74.9} &\textbf{69.3} &61.3 &60.3 \\
		\hline
		\textit{FPConv} (ours) & 68.7&  \textbf{94.8}&	\textbf{97.5}&	82.6&	42.8&	41.8&	58.6&	73.4&	\textbf{71.0}&	\textbf{81.0}&	59.8&	61.9&	\textbf{64.2}&	\textbf{64.2} \\
		\bottomrule
	\end{tabular}
	\caption{Detailed semantic segmentation scores on S3DIS 6-fold cross validation.}
	\label{table:s3dis_6fold}
\end{table*}

\end{document}